\documentclass[runningheads]{llncs}
\usepackage{graphicx}
\usepackage[caption=false]{subfig}
\usepackage{color}

\begin{document}
\title{A Hierarchical Approach for Visual Storytelling Using Image Description}
%
%
\author{Md Sultan Al Nahian \and
Tasmia Tasrin \and
Sagar Gandhi \and
Ryan Gaines \and
Brent Harrison}
%
%
\institute{Department of Computer Science, University of Kentucky
\email{\{sa.nahian,tasmia.tasrin,sga267,ryan.gaines,brent.harrison\}@uky.edu}}
\maketitle              
\begin{abstract}
One of the primary challenges of visual storytelling is developing techniques that can maintain the context of the story over long event sequences to generate human-like stories. 
  In this paper, we propose a hierarchical deep learning architecture based on encoder-decoder networks to address this problem. 
  To better help our network maintain this context while also generating long and diverse sentences, we incorporate natural language image descriptions along with the images themselves to generate each story sentence. 
  We evaluate our system on the Visual Storytelling (VIST) dataset~\cite{huang2016visual} and show that our method outperforms state-of-the-art techniques on a suite of different automatic evaluation metrics.
  The empirical results from this evaluation demonstrate the necessities of different components of our proposed architecture and shows the effectiveness of the architecture for visual storytelling.
\end{abstract}
\section{Introduction}

Computational storytelling is the task of automatically generating cohesive language that describes a sequence of correlated events or actions. 
Prior work on computational storytelling has mainly focused on plan-based approaches for generating narratives~\cite{ywcr:plan}. 
Planning based approaches often rely on complex domain models that outline the rules of the world, the actors involved, and the actions that each actor can take. 
This type of story generation, often called closed-world storytelling, is able to generate coherent stories, but are restricted in the types of stories they can generate by the domain model. 

Recently there has been an increased interest in {\em open-world story generation}. 
Open-world story generation refers to generating stories about any domain without prior knowledge engineering and planing~\cite{lpxwsbm:event}.
With the increased effectiveness and sophistication of deep learning techniques, deep neural networks, such as sequence-to-sequence networks, have been shown to be effective in open-world story generation. 
The primary advantage that these techniques have over planning systems is that they do not require extensive domain modeling to be effective. 
This makes them an effective tool for open-world story generation. 

{\em Visual storytelling} is an extension to the computational storytelling problem in which a system learns to generate coherent stories based on a sequence of images. 
Visual storytelling is a more challenging problem because the sentences need to be not only cohesive, but also need to consider both the local context of images and the global context of whole image sequence. 
There have been recent successes in generating natural language that is conditioned on images. 
These successes are primary in tasks such as image captioning~\cite{vtbe:showtell,densecap,kjkf:description}.
Visual storytelling presents a different challenge from image captioning in that the language generated is often more abstract, evaluative, and conversational~\cite{huang2016visual}. 
In addition, techniques need to identify and understand the relations among the scenes of the images and describe them through logically ordered sentences. 
The task also needs to consider the completeness of the story. 

There have been successes in visual storytelling, however.
Approaches utilizing deep learning have proven, overall, to be effective at this task~\cite{hrnn,rico:context}.
Though these approaches have achieved competitive results, the stories they produce are often comprised of short sentences with repeated phrases. 
In some cases, the generated sentences fail to tell a coherent story and in some cases they fail to capture image contexts. 


One of the primary challenges in both computational storytelling and visual storytelling is determining how to maintain story context for long event sequences. 
In this paper, we address this challenge with techniques inspired by how humans usually form stories.
To construct a story, a human needs to form a plot for the story at first. 
In visual storytelling, this can be done by going through all the images and extracting the key context from them to form the premise of the story. 
After that, the sentences are made by going through the images one by one.
In order to ensure the coherency in the story, we need to articulate the temporal dependencies among the events of the images. 
This can be achieved by summarizing the events generated in the previous sentences and considering the summary during making the current sentences of the story.
For instance,~\cite{slm:components} use the sentence generated for the previous image to generate the sentence for the current image.
This approach can have a cascading error effect in which errors can compound if the quality of the previously generated sentences is bad. 
As a result, over time the context of the story can drift from the original context. 

To emphasize logical order among the generated sentences and to help our network architecture better maintain context over time, we have passed image descriptions into the network along with the images themselves. 
This helps to minimize the effect of cascading error as mentioned above.
The descriptions are more specific and literal statements about the content of the images. 
The intuition behind adding these descriptions is that they can aid the network in understanding the context of the image and help the network to extract the flow of the events from the subsequent images.

To evaluate our system, we examine its performance on the Visual Storytelling (VIST) dataset using automatic evaluation metrics BLEU, CIDEr, METEOR and, ROUGE\_L.
Using these evaluation metrics, we have demonstrated how well our proposed architecture can learn description and image context and combine both of them to create human-like coherent visual narratives. 

The major contributions of our work can be summarized as follows:
\begin{itemize}
\item An end-to-end hierarchical deep neural network to generate open story from visual content. 
\item Exploration into the use of natural language image descriptions in visual storytelling. 
\item An evaluation of our architecture on a large corpus of visual storytelling data against state-of-the-art deep learning techniques.
\end{itemize}

\section{Related Work}
Research on computational story generation has been explored in two ways: Closed-world story generation and open-world story generation. 
Closed-world story generation typically involves the use of predefined domain models that enable techniques such as planning to be used to generate the story. 
Open-world story generation involves automatically learning a domain model and using it to generate stories without the need for relearning or retraining the model~\cite{bcm:mark}. Due to their ability to reason over sequences of textual input, sequence-to-sequence networks are typically used to perform open-world story generation. 
To better help these networks maintain context over long story sequences, many researchers have chosen to make use of event representations that distill essential information from natural language sentences~\cite{bcm:mark,lpxwsbm:event,planw}. These {\em event representations} make story generation easier in that the network only needs to focus on generating essential information. 
In our work, we perform the more complex task of reasoning over both story information as well as visual information.
In addition, we do not make use of event representations, choosing instead to generate full text sentences. 


Visual narrative has been explored previously, primarily utilizing planning-based approaches~\cite{cr:plot}.
With the release of the first large-scale, publicly available dataset for visual storytelling~\cite{huang2016visual}, approaches based on machine learning have become more viable for the task.
In~\cite{huang2016visual}, they propose a  sequence-to-sequence network to generate story from image sequence which has been being used as a strong baseline for the visual storytelling task.
\cite{src:pipe} has proposed a visual storytelling pipeline for task modules which can serve as a preliminary design for building a creative visual storyteller.
~\cite{slm:components} has proposed a visual storytelling system where previous sentence is used to generate current sentence. \cite{kim:glac} and \cite{wang:arel} are the two winners from the VIST challenge in 2018. GLAC Net~\cite{kim:glac} generates visual stories by combining global-local attention and provides coherency to the stories by cascading the information of the previous sentence to the next sentence serially. 
In \cite{wang:arel}, an adversarial reward learning scheme has been proposed by enforcing a policy model and a reward model. 
One common limitation of these approaches is that the models often generate short sentences and are prone to repeating certain phrases in their story sentences. 
We believe that by utilizing image descriptions we are bootstrapping the language learning process.
This enables our method to produce more diverse, human-like sentences that are longer than the ones generated by previous approaches while still maintaining coherence. 

\section{Methodology}

\begin{figure}
\centering
\includegraphics[scale=0.80]{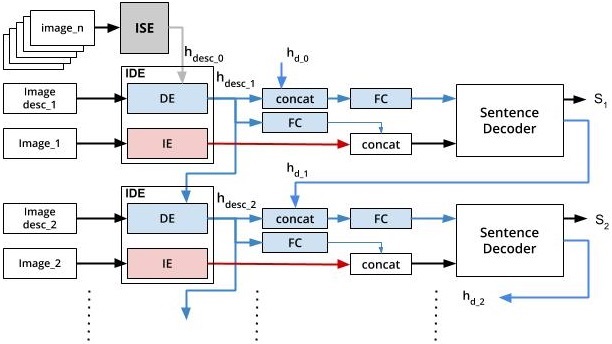}
\caption{The Proposed Hierarchical Context Based Network. Image Sequence Encoder(ISE) takes all the images and encode them to create a vector. Image-Description Encoder(IDE) is composed of two components: Description Encoder(DE) to encode the description and Image Encoder(IE) to encode the image. In each iteration, the sentence decoder(SD) generates a sentence, word by word, conditioned on the vectors coming from DE and IE. In the figure, two iterations have been shown.} \label{hcbnet}
\end{figure}

In this work, we propose a hierarchical encoder-decoder architecture which we call a {\em Hierarchical Context-based Network (\textbf{HCBNet})} to generate coherent stories from a sequence of images.
Fig.~\ref{hcbnet} shows the overview of our architecture. 
The network has two main components: 1. A hierarchical encoder network that consists of two levels and 2. a sentence decoder network.
The first level of the hierarchical encoder, referred to as the {\em Image Sequence Encoder (ISE)}, is used to encode all the images of the sequence and create a single encoded vector of the image sequence.
In the next level, there is a composite encoder, referred to as the {\em Image-Description Encoder(IDE)}. 
It takes in two inputs: an image and description of that image. 
The IDE consists of two encoders: an {\em Image Encoder} that is used to encode the individual image and a {\em Description Encoder} that is used to encode the description of the image in each time step.
After each iteration, the decoder network generates one sentence, word by word, of the story as output. 
The initial state of the first time step of the Description Encoder comes from the Image Sequence Encoder as shown by the grey arrow in Fig. 1.
Each of the components of our proposed architecture will be discussed further below.

\subsection{Hierarchical Encoder Network}
\begin{figure}[t!]
\centering
\includegraphics[scale=0.70]{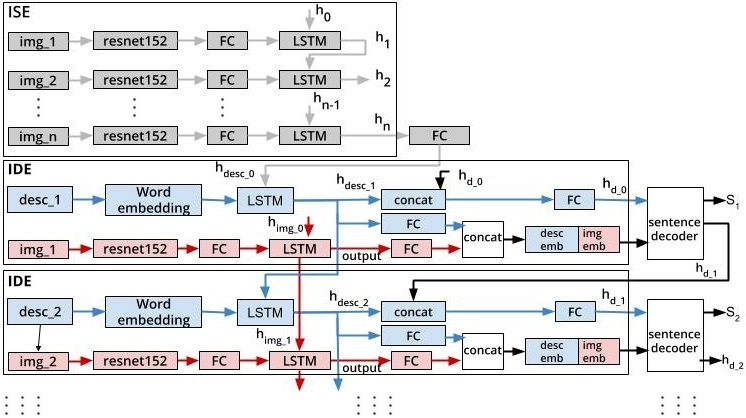}
\caption{Detailed Architecture of the Encoder. ISE (represented by grey color) generates sequence embedding vector which is used as the initial hidden vector of DE in the first iteration of IDE. IDE produces two vectors: initial hidden vector and input of the sentence decoder. Blue color represents the DE network and red color represents the IE network. } \label{encoder}
\end{figure}

As mentioned earlier, the Hierarchical Encoder of our architecture is composed of two levels of encoder: the Image Sequence Encoder(ISE) and the Image-Description Encoder(IDE). 
The IDE is, itself, composed of two different encoders that are tasked with identifying different types of context in our input. 
Fig. \ref{encoder} presents a detailed representation of the hierarchical encoder.

\subsection{Image Sequence Encoder}
This is the first encoder which is meant to help the network understand the high-level context of the story based on the images that it has seen thus far. 
It takes an image sequence as input and in each time step, uses a convolutional neural network (CNN) to embed each individual image into a single vector. 
This vector is passed to an additional fully connected (FC) layer and then to LSTM network \cite{hs:lstm}. 
The output hidden state of the LSTM is forwarded to the LSTM of the next iteration. 
For the CNN, we have used pretrained model. 
We have experimented with pretrained VGG19~\cite{Simonyan14c} and resnet152~\cite{He:resnet152} to extract features from the images.

After the iteration has been finished for each of the images in the sequence, we take the final hidden state of the LSTM unit and pass it to a FC layer. 
The output vector of the FC layer, referred to as the sequence embedding vector, represents the global context of the image sequence which works as the premise of the story. 
This sequence embedding vector is used as the initial hidden vector of the Description Encoder (shown in Fig. \ref{encoder}). 
Therefore the global context of the story passes through the network through time. 
This helps the network to better understand and maintain the theme of the story throughout the entire sequence. 

\subsubsection{Image-Description Encoder}
The Image-Description Encoder (IDE) takes an image and corresponding description as input in each time step. 
In the first time step, it takes the sequence embedding vector from the ISE as initial hidden state as mentioned above. 
The IDE is a composite encoder with two modules: The first is the \textit{Image Encoder (IE)}, which is used to extract the context vector from an individual image. 
The second module is the \textit{Description Encoder (DE)}, which is used to extract the context vector from the corresponding image description.

\paragraph{Image Encoder}
This component is used to deduce the context of the current image given additional information about the previous images in the sequence.
The current image is sent to a Convolutional Neural Network(CNN) pretrained using the resnet152 model for feature extraction. 
The extracted image feature vector is passed through to a FC layer and a recurrent neural network. 
Fig.~\ref{encoder} outlines the overview of the encoder with red arrows. 
It shows that the LSTM network takes a hidden state as input as well. 
This is the feature vector from the Image Encoder of the previous time step. 
The LSTM of current time step also generates a hidden state and an output vector.
We pass the hidden state to the Image Encoder of the next step and pass the output to an FC layer to form the image embedding vector.
The hidden states of Image Encoder propagate the local image context from one time step to the next. 

\paragraph{Description Encoder}

The Description Encoder is used to extract information about the current description and to reason over information contained in previous image descriptions.
We use the image description to help maintain temporal dependencies between the sentences.
Before passing the description into the RNN(LSTM), we preprocess the description and pass it through to an embedding layer. 
The final hidden state of the LSTM contains contextual information about the image description. 
This can be thought of as \textit{theme} type information that is used to condition the output of our network. 
It is passed to the Description Encoder of the next iteration where it will be used as the initial state for the LSTM (shown as the blue connection in Fig. \ref{encoder}). 
By doing this, we help ensure that the context of the current sentence is passed to the next iteration. 

The hidden state of the DE is also passed to a FC layer to form a vector referred as the description embedding. The description embedding and image embedding are concatenated and forwarded to the Sentence Decoder.
The hidden state of the DE is concatenated with the final hidden state of the decoder of previous time step as well. A FC layer is applied on the concatenated vector and forwarded to the decoder to form the initial hidden state of the Sentence Decoder(SD). The relations are demonstrated in eqn. \ref{eq:decod_hid_0}:

\begin{equation} \label{eq:decod_hid_0}
decod\_hid_{s,0} = FC(concat(desc\_hid_s, decod\_hid_{s-1,t}))
\end{equation}
\begin{equation}\label{eq:decod_hid_t}
decod\_hid_{s,t} = decod\_hid_{s,t-1}
\end{equation}

From the hidden state of the previous sentence decoder, the network receives information about what has been generated before and the hidden state of the description provides the theme of the current sentence. 
This information enables the network to construct an image specific sentence while maintaining the flow of the events across the sentences.

\begin{figure}
\centering
\includegraphics[scale=0.7]{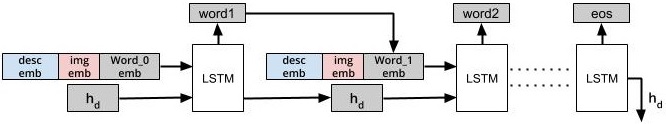}
\caption{Sentence Decoder}
\label{fig:decoder}
\end{figure}

\subsubsection{Sentence Decoder}
The Sentence Decoder (SD) uses a LSTM to generate the text of the visual narrative (seen in Fig.~\ref{fig:decoder}).
This LSTM network uses the contexts generated by the IDE to construct a story events word by word.
In the beginning of a sentence, the initial hidden state of the sentence decoder is formed by eqn. \ref{eq:decod_hid_0}. Then it propagates the hidden state of current time step to next time step of a sequence (represented in eqn. \ref{eq:decod_hid_t}).
Input of the SD is formed by concatenating the description embedding, image embedding, and word embedding of the previous word (eqn \ref{eq:sent_dec_in}). This process is repeated in every word generation step of a sentence. It works as ``hard attention'' on image and description context. From the image embedding vector, the decoder gets information on the local features of the current image, while the description embedding provides both the overall story context and image specific context to the decoder.

\begin{equation} \label{eq:desc_emb}
desc\_embed_s = FC(desc\_encoder\_hidden\_state_s)
\end{equation}
\begin{equation} \label{eq:sent_dec_in}
sent\_decod\_input_{s,w} = concat(desc\_embed_s, img\_embed_s, word\_embed_{s, w-1})    
\end{equation}


\section{Experimental Setup}
In this section, we give the details about our experimental methodology and the implementation that we used for testing.
First, we describe the dataset used in the experiment followed by how we chose to preprocess the data. 
Then we will discuss information about the network parameters used during testing.
Finally, We give an overview about the baseline architectures against which the proposed architecture has been compared. 
Afterward, we discuss about the evaluation metric and evaluation criteria. 

\subsection{Dataset}
To evaluate our architectures, we have used the Visual Storytelling Dataset (VIST) \cite{huang2016visual} which is a sequential vision-to-language dataset.
This dataset consists of image sequences and associated stories for each sequence.
Each image sequence consists of 5 images and, therefore, five story sentences. 
In addition to this information, some images have an image description associated with them which is referred as ``Descriptions of images-in-isolation'' in the VIST paper. 
In this paper, we choose to use the term ``image description" instead of ``Descriptions of images-in-isolation".
The dataset, in total, contains 40155 stories in the training set, 4990 stories in the validation set and, 5055 stories in the test set.
It contains 154430 unique images in the training set, 21048 images in the validation set and 30000 images in the testing set.

Recall that our approach makes use of image descriptions to help generate narratives. 
Some images, however, do not have an associated description. 
Images that did not have an associated description were discarded, as were any stories that contained images that did not have an associated description.
This process has reduced the total number of training stories to 26905, validation stories to 3354 and, test stories to 3385.


\subsection{Preprocessing}
In the preprocessing step, we have corrected the misspellings from the image descriptions and story texts. 
Stop words have been removed from the image descriptions, but not from the story texts.
To build the vocabulary, we have taken all the words which have been appeared at least three times in the story text, which results in a vocabulary of size 12985.
Words that appear below this threshold are replaced with a symbol representing an unknown word.
We have used pretrained resnet152 and VGG19 models to encode the images. As these two models take input of size 224X224, we have resized all of the training, validation and test images to the required size.

\subsection{Experimented Architectures}
We have evaluated our proposed method, HCBNet, against two of the state-of-the-art architectures for visual storytelling: AREL~\cite{wang:arel} and GLAC Net~\cite{kim:glac}. 
We have trained these two networks from scratch with the same dataset used to train our proposed model. 
To evaluate the need for different aspects of HCBNet, we also perform an ablation study. 
In this ablation study, we test four different versions of HCBNet. They are:
\begin{itemize}
    \item \textbf{HCBNet:} This is the standard HCBNet demonstrated in Fig. \ref{hcbnet}.
    \item \textbf{HCBNet without previous sentence attention:}
    In this version, the final hidden state of an iteration of the sentence decoder is not used to form the initial hidden state of the next iteration of the sentence decoder. The hidden state of the description encoder is used as the initial hidden state of the sentence decoder. The sentence decoder gets the context of the previous sentence from the hidden state of the description encoder.
    \item \textbf{HCBNet without description attention:}
    In this version, previous sentence attention is used in the sentence decoder, but the description embedding is not passed to form the input of the sentence decoder. Here, the sentence decoder does not use attention on the image description during each step of an iteration. It gets the description information from the hidden state of the description encoder in the beginning of an iteration.
    \item \textbf{HCBNet using VGG19:} In all of our experiments, we have used pretrained resnet152 as the CNN. Here, we have experimented the HCBNet with pretrained VGG19 model to check if there is any significant performance difference when using a different pretrained CNN.
\end{itemize}
\subsection{Network Parameters}
All of the networks in the experiment have been trained using same parameters. 
The learning rate is 0.001 with learning rate decay 1e-5.
The batch size is 36, and the vocabulary size is 12985. 
The size of the LSTM units are 1024 which is same for all of the versions of HCBNet. 
We have used a multilayer LSTM with 2 layers. 
To prevent the overfitting, we have used batch normalization and dropout layers. 
Batch normalization has been applied on each FC layer that is followed by an LSTM cell in both the ISE and IE module. Another batch normalization has been used on the final output vector (FC layer) of the ISE module. 
We have applied dropout at a rate of 0.5 on each LSTM cell as well as the output vector of the LSTM cell of the SD. 
The training process is stopped when the validation loss no longer improves for 5 consecutive epochs (25 to 32 epochs in these experiments). 
We have used Adam optimizer to optimize the loss.

\begin{table}
\caption{Automatic evaluation metrics results from the experiments}\label{table:automated_metrics}
\begin{tabular}{ |p{2.5cm}|p{1.3cm}|p{1.3cm}|p{1.3cm}|p{1.3cm}|p{1.3cm}|p{1.55cm}|p{1.55cm}|}
 \hline
  & BLEU-1 & BLEU-2 & BLEU-3 & BLEU-4 & CIDEr & METEOR & ROUGE\_L\\
 \hline
 AREL(baseline) & 0.536 & 0.315 & 0.173 & 0.099 & 0.038 & 0.33 & \textbf{0.286}\\ 
 \hline
 GLAC Net(baseline) & 0.568 & 0.321 & 0.171 & 0.091 & 0.041 & 0.329 & 0.264\\
 \hline 
 HCBNet & 0.593 & \textbf{0.348} & 0.191 & 0.105 & 0.051 & \textbf{0.34} & \textbf{0.274}\\
 \hline
 HCBNet(without prev. sent. attention) & \textbf{0.598} & 0.338 & 0.180 & 0.097 & \textbf{0.057} & 0.332 & 0.271\\ 
 \hline
 HCBNet(without description attention) & 0.584 & 0.345 & \textbf{0.194} & \textbf{0.108} & 0.043 & 0.337 & 0.271\\ 
 \hline
 HCBNet(VGG19) & 0.591 & 0.34 & 0.186 & 0.104 & 0.051 & 0.334 & 0.269 \\
 \hline
\end{tabular}
\end{table}

\subsection{Evaluation Metrics}

In order to evaluate our proposed architecture, we have used the following automatic evaluation metrics: BLEU~\cite{bleu}, METEOR~\cite{meteor}, ROUGE\_L~\cite{rouge} and CIDEr~\cite{cider}. 
BLEU and METEOR are machine translation evaluation metrics that are widely used to evaluate story generation systems. 
To use BLEU, one must specify an n-gram size to measure.
In this paper, we report results for 1-grams, 2-grams, 3-grams, and 4-grams.
ROUGE\_L is a recall based evaluation metric primarily used for summarization evaluation.
CIDEr is different from translation metrics in that it is a consensus based evaluation metric. It is capable of capturing consensus and is, therefore, able to better to evaluate ``human-likeness'' in the story than metrics such as BLEU, METEOR or ROUGE\_L. 



\section{Results and Discussion}
We evaluate the performance of our proposed method HCBNet against the baseline networks AREL and GLAC Net by using the automated evaluation metrics BLEU, CIDEr, METEOR and ROUGE\_L. 
The scores are shown in the Table \ref{table:automated_metrics}. 
These results demonstrate that HCBNet outperforms the GLAC Net on all of the metrics and AREL on all of the metrics except for ROUGE\_L. 
Though AREL performs better than HCBNet in ROUGE\_L, inspection of the stories generated by each network indicates that the quality of the stories generated by HCBNet are higher than those generated by AREL.

It is important to note that these metrics on their own do not necessarily indicate that our method produces interesting, or even coherent, stories. 
Recall that we claimed earlier that the stories produced by GLAC Net and AREL often suffer from having short sentences with repeated phrases. 
One of our hypotheses in this paper is that utilizing image description information should enable us to generate stories with longer and more diverse sentences. 
We perform an analysis to provide some intuition on whether this is the case. 
Specifically, we compare the average number of words per sentence and the number of unique 1, 2, 3, and 4-grams generated by each network. 
From Table \ref{table:ngrams}, we can see that the average number of words per sentence is highest for AREL among the three networks. 
But number of unique 1-grams is only 357 for AREL, where HCBNet has 1034 unique 1-grams. 
This behavior is consistent across all n-grams tested.
This provides support to our claim that these baselines tend to generate repetitive phrases and provides support to our claim that HCBNet can produce more diverse sentences. 
Interestingly enough, this also could explain why AREL performed well on ROUGE\_L. 
ROUGE\_L is meant to measure a model's recall on a reference sentence, which is likely to be high if one produces short sentences. 

As shown in the Table \ref{table:automated_metrics}, the CIDEr score of HCBNet is higher than GLAC Net and AREL.
This indicates that our  model has a greater ability to generate ``human-like'' stories than compared to AREL and GLAC Net. 
It is also notable that we see the greatest difference between our network and our baselines through this metric. 
We feel that this, especially when combined with the results outlined in Table~\ref{table:ngrams}, further indicate that our network produces stories that are more diverse and, potentially, human-like while still maintaining story context. 

As mentioned in section 4.3, we have also experimented with three other versions of HCBNet to see the effectiveness of different components of the network. 
HCBNet without previous sentence attention gives higher score in CIDEr and slightly better score in BLEU-1, but adding the same component into the network significantly increases the score of other metrics. 
HCBNet without using description attention performs poorly for CIDEr. Incorporating the  description embedding into the input of sentence decoder not only improves the score of CIDEr remarkably but also METEOR, ROUGE\_L, BLEU-1 and BLEU-2 scores. 
We believe that this indicates that the description attention helps the network resist context drift and helps keep the story cohesive. 

\begin{table*}[t!]
\centering
\caption{Experiment Results based on Word properties}\label{table:ngrams}
\begin{tabular}{ |p{2cm}|p{1.6cm}|p{1.6cm}|p{1.4cm}|p{1.4cm}|p{1.4cm}|p{1.5cm}|p{1.5cm}|}
 \hline
  & avg. no. of words per story & avg. no of words per sent. & 1-gram & 2-gram & 3-gram & 4-gram \\
 \hline
 AREL & 29.893 & 7.03 & 357 & 924 & 1526 & 1979\\ 
 \hline
 GLAC Net & 29.826 & 5.996 & 837 & 2586 & 4069 & 4628\\
 \hline 
 HCBNet & \textbf{30.7} & \textbf{6.141} & \textbf{1034} & \textbf{3324} & \textbf{5292} & \textbf{5966}\\
 \hline
\end{tabular}
\end{table*}

\begin{table}
\caption{Comparison of the stories among the networks}\label{table:story}
\begin{tabular}{ |p{1.8cm}|p{10.8cm}|}
 \hline
 \multicolumn{2}{|c|}{ \includegraphics[scale=0.90]{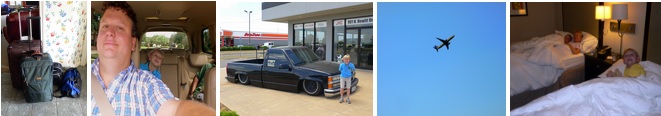}}\\
 \hline
 Preprocessed Descriptions & 1. suitcase stand near wall flower 2. man seat belt front boy back seat 3. boy stand front lowered black park truck 4. lone jumbo plane fly sky 5. old woman child lay hotel room\\
 \hline
 Ground Truth Story & everybody 's packed for the trip. kids are in the car and ready to go. we dropped the rental off and [male] got a picture next to the car. [male] asked us if this was our plane. we spent the night in the hotel after a long day of travel.\\
 \hline
 AREL & i went to the bar yesterday. the kids were so happy to be there. we had a great time at the park. the \textless UNK\textgreater was very good. after the party, we all had a great time.\\
 \hline
 GLAC Net & i went to the beach yesterday. there were a lot of people there. i had a great time. it was a beautiful day. afterward we all got together for a group photo.\\
 \hline
 HCBNet & the family went to a farm for their vacation. they got a little tired and took pictures of each other. they were able to get a picture of a tractor. then they saw a plane. after that, they took a break outside.\\
 \hline
\end{tabular}
\end{table}

\begin{table}
\caption{Example stories generated by HCBNet}\label{table:example_story}

\begin{tabular}{ |p{12.5cm}|}
 \hline
  \includegraphics[scale=0.88]{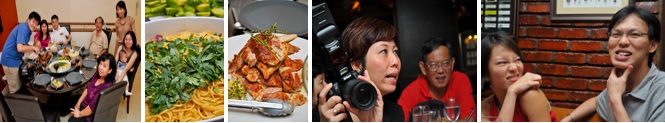}\\
 \hline
 the family gathered for a special dinner. they had a lot of food and drinks. there was also a lobster dish. [female] was happy to be there. she was so excited to see her friends.\\
 \hline
 \includegraphics[scale=0.81]{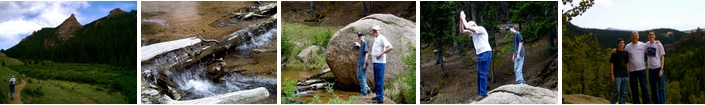}\\
 \hline
 the family went on a hike to the mountains. we saw a beautiful waterfall. it was a nice day for a walk. he was very excited. after that we took a picture of him.\\
 \hline
\end{tabular}
\end{table}

To provide a better understand and highlight the differences between the stories generated by our network and by our baselines, we have provided some illustrative examples of stories generated by each network.
Table~\ref{table:story} shows a comparison between the story generated by our network and the baseline networks. 
The first thing to note is that the stories generated by the baseline networks are relatively vague and rely on general phrases about having a great time.
In addition, they often disregard the context that each image provides.
If we examine the story generated by HCBNet, we can see that our network correctly interprets the theme of the story as ``vacation''. 
The corresponding generated sentence of image 3 is interesting, though. 
In the image, though the vehicle is a truck, HCBNet describes it as a tractor. We believe this is because the network correctly identifies a vehicle, but wants to remain consistent with the fact that it says the family is visiting a farm in the first sentence. 
We feel that this type of behavior shows our network is able to balance maintaining story context along with maintaining image context.
In the last sentence, though it believes the people to be outside, it understands that people are taking a break. 
We have also provided more examples of the stories that our network can generate in Table \ref{table:example_story}.
We feel that these results combined with the results achieved on our automatic evaluation metrics provide significant evidence for our claim that HCBNet can produce high quality visual narratives.

\section{Conclusion and Future Work}
In this paper, we introduce HCBNet, a hierarchical, context-based neural network that incorporates image description data for performing visual storytelling.
In addition, we evaluate our approach using a variety of automatic evaluation metrics and show that HCBNet outperforms two state-of-the-art baselines.
Our results indicate that our proposed architecture is able to learn the expected flow of events conditioned on the input images and use this knowledge to produce a cohesive story. 
As our future work, we plan to expand our evaluation to include a human subjects study so that we can explore how humans perceive the stories generated by our system.

%
%
%
 \bibliographystyle{splncs04}
 \bibliography{samplepaper}
\end{document}